\documentclass{IEEEtran}
\usepackage{graphicx}
\usepackage[utf8]{inputenc}
\usepackage{cite}
\begin{document}
\title{MOBA: a New Arena for Game AI}

\author{%
Victor do Nascimento Silva$^{1}$ and Luiz Chaimowicz$^{2}$%
\thanks{$^{1}$V. N. Silva is with the Department of Computing Science, University of Alberta, Canada
        {\tt\small vsilva@ualberta.ca}}%
\thanks{$^{2}$Luiz Chaimowicz is with the Department of Computing Science, Federal University of Minas Gerais, Brazil.
        {\tt\small chaimo@dcc.ufmg.br}}%
\thanks{*This work was partially supported by FAPEMIG and CNPq}
}
\maketitle

\begin{abstract}

Games have always been popular testbeds for Artificial Intelligence (AI). In the last decade, we have seen the rise of the Multiple Online Battle Arena (MOBA) games, which are the most played games nowadays. In spite of this, there are few works that explore MOBA as a testbed for AI Research. In this paper we present and discuss the main features and opportunities offered by MOBA games to Game AI Research. We describe the various challenges faced along the game and also propose a discrete model that can be used to better understand and explore the game. With this, we aim to encourage the use of MOBA as a novel research platform for Game AI.  


\end{abstract}


\section{Introduction}

From Arthur Samuel's research on Checkers~\cite{samuel1959}, through Deep Blue's efforts in defeating a human chess champion~\cite{campbell2002} to the recent accomplishments of AlphaGo~\cite{silver2016}, games have always been an important drive for Artificial Intelligence (AI). Games provide a controlled test environment with interesting challenges, where novel AI techniques can be tested. Moreover, playing competitively against humans has always motivated AI researchers and has proved to be difficult in most games.


In the last decade, Real Time Strategy (RTS) Games have emerged as a challenging testbed for AI~\cite{buro2003}. More than just rational behavior, these games require agents that reason and act as fast as possible in partially known environments. In RTS Games, AI techniques have to deal with a broad range of variables, that could influence the chances of winning and losing the game. Real-time games also offer a wider range of actions to the player, and, sometimes, a large number of agents to be controlled~\cite{hagelback2012}.

One of the most successful testbeds in this context is the RTS Game Starcraft. Specially, the Brood War version of Starcraft became very popular among researchers, being used in many works and competitions~\cite{buro2012}. It requires abilities in different areas such as unit coordination, resource management and combat skills and presents several challenges to AI agents that have to deal with a complex state space, feature analysis and some kind of intuition involved in Starcraft gameplay. Due to this, humans has historically performed better than computers in Starcraft \cite{ontanon2013, Ackerman2015}. 


We believe that Starcraft has become a standard research platform because of two main factors: a) the availability of a programming environment that allowed researchers to develop their agents \cite{chung2005}; and b) the large player base of the game, allowing researchers to compare their agents with humans. These factors make the research easier, and makes it more relevant, because it impacts more people. Moreover, it is better to perform user test in an environment that players are familiar with, because they feel comfortable, and already know how to behave in the game. 

Recently, the game community has witnessed the rise of MOBA (Multiple Online Battle Arena) Games. With titles such as League of Legends, Smite and Dota 2, MOBA games are responsible for almost 30\% of the online gameplay around the world~\cite{murphy2015}. League of Legends alone consumes approximately 23\% of this gameplay, and became the most played game of the world, surpassing the MMORPG World of Warcraft~\cite{gaudiosi2012}. 

MOBA can be considered a sub-genre of the RTS, inheriting some of its characteristics. However, instead of focusing on the ability of maneuvering large amounts of units and building fortresses, a set of actions known as Macromanagement, MOBAs have a strong focus on Micromanagement which consists in mastering a set of actions and their best use in the control of a small set of units. In MOBA games, this is generally known as mechanics, and players with these fine abilities normally excel in the game.

Due to their specific gameplay, MOBA games present a different and interesting set of challenges for Game AI. But, in spite of their huge commercial success, they have not yet been extensively explored as a testbed for game AI such as their RTS ancestors. Thus, in this paper we present and discuss the main challenges and opportunities offered by MOBA games to Game AI Research. We break down this discussion considering the specific phases of the gameplay and also in more general terms related to game mechanics. 
Moreover, we present a general model based on a MOBA abstraction that can facilitate related research in more general settings such as autonomous agents, computational intelligence and learning.


The remainder of this paper is organized as follows: we give an overview of the MOBA genre in Section \ref{sec:MOBA}. We follow by discussing the MOBA phases and their specific challenges in Section \ref{sec:phases}. The mechanic challenges found in MOBA are described in Section \ref{sec:mechanic} while the Discrete Model proposed for the MOBA environment is presented in \ref{sec:discrete}. Finally, we present the related research that have been done using MOBA in Section \ref{sec:related} and discuss the conclusions and future work in Section \ref{sec:conclusion}.

\section{Multiplayer Online Battle Arena}
\label{sec:MOBA}

The MOBA genre has its roots on the RTS genre, and even shared game platforms in the early days. Instead of focusing on the management of an army, exploring maps and build a fortress, MOBA focuses on the finest mechanic abilities, such as combos, kiting, abilities, among others.

The core gameplay of most MOBAs is the control of a single unit, commonly called hero. This unit has a set of special powers and characteristics that allow it to combat other units. Each player receives his chosen hero in the level 1, and should evolve his hero to the max level offered by the game. Thus, each game is different from another, and the environment is not persistent. The main goal is to conquer a single structure located at the opposite side of the map.

Another important characteristic of MOBA is that it is not played alone. The common game mode of most MOBAs requires two teams of five players. Each player controls a single hero that should assume a role in the team, composing the team's strategy. The role consists of the function that the hero will assume in the match, e.g. Carry, Initiator, Tank, Ganker, among others. Each team has a base, which contains a structure that should be conquered by the opponent team. To do so, the team should fight against strong defensive structures, called turrets. 

\begin{figure}[h]
\begin{center}
\includegraphics[scale=0.26]{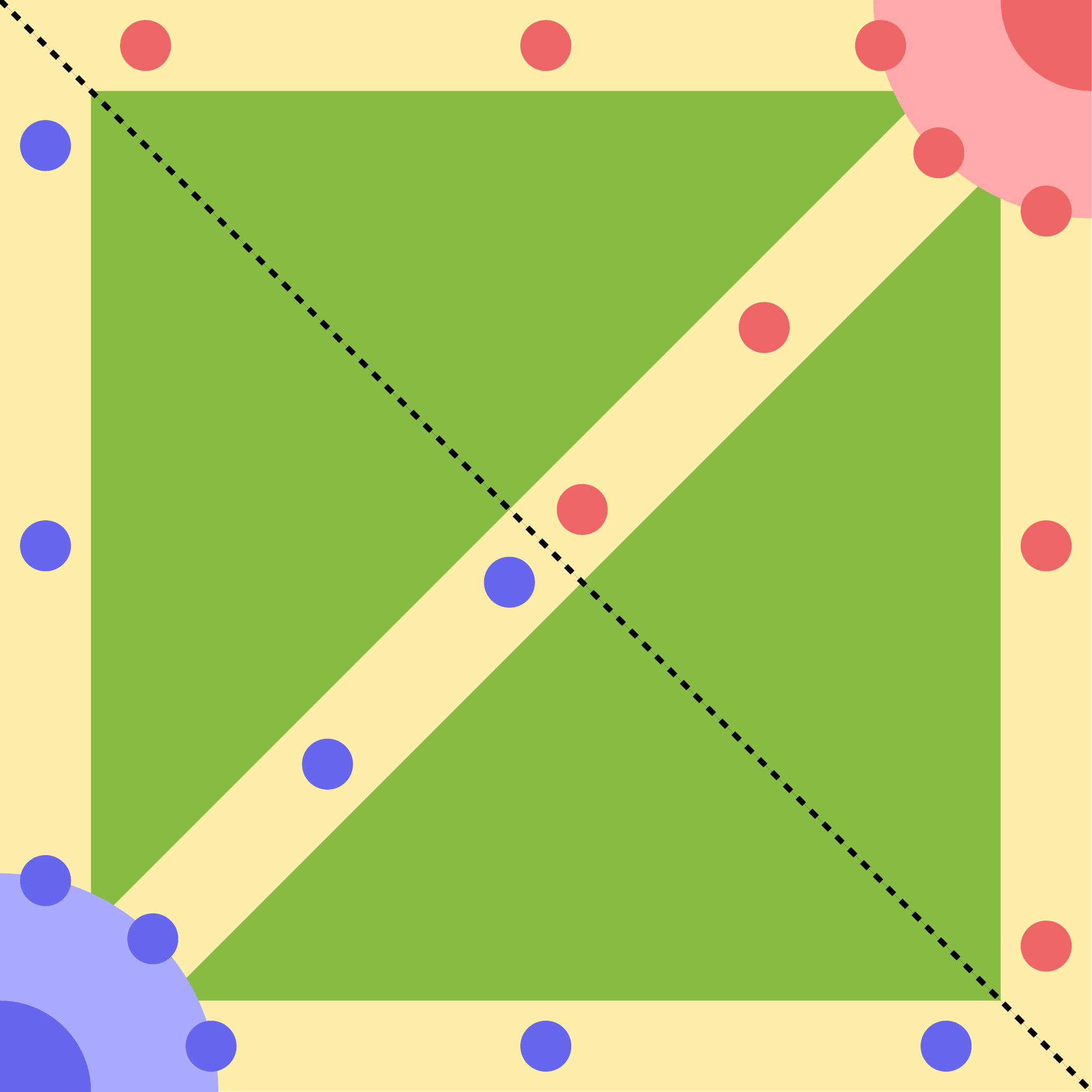}
\caption{A generic map found on MOBA games. Red shapes represent a team while blue represent another. Circles represent turrets. The green area is the jungle and the yellow area are the lanes. The corner areas are the team's base.}
\label{fig:MOBAMap}
\end{center}
\end{figure}

A general MOBA map is composed by three lanes, and each lane can contain up to three turrets. To conquer a turret closer to the base, the team should firstly destroy the precedent turret in the same lane. A view of a generic MOBA map can be seen in Figure \ref{fig:MOBAMap}.

An important part of RTS gameplay is to collect resources from specific places to build structures and units. For instance, in Starcraft this resource collection is done in sources of gas and crystals, that are later used to create units and structures. In the MOBA gameplay, the resource collection task is also present, but with different characteristics. Instead of simply sending units to collect the materials from a source, MOBA rely on the player skills to perform the so called {\em farming}. Farming consists in killing certain units known as {\em creeps} or {\em minions} in order to get gold. These units spawn in waves from each team base, and follow a lane fighting their way to the enemy base. Besides, it is possible to obtain gold from neutral minions, which live in the jungle, the free space between the lanes. Lastly, killing enemy heroes or destroying enemy structures also give the hero or the team a certain amount of gold. Therefore, more mechanically skilled players can collect more resources than the least skilled ones. We can consider that collecting resources in MOBA is a harder task than in RTS since it involves skills like positioning, skill/attack control and target selection, while RTS requires the player to simply send units to a specific place to collect resources.

The gold in MOBAs is not used to buy structures or units. Instead, it is used to buy items for the controlled hero. These items improve the hero abilities and status, making it more powerful. Besides, there are items that give the hero special powers and can be used during team fights or to disable enemies. Thus, instead of focusing in building a fortress, MOBA requires the player to improve his/her own hero, preparing it to succeed in the game. 

Lastly, MOBA focuses on team cooperation to defeat an enemy team. Differently from RTS, which values players that are able to manage many units, explore the map and build structures, MOBA rewards players with higher mechanical skills that are also capable of cooperating with other players to execute a successful strategy. Therefore, besides individual abilities, cooperation is an important feature in MOBA gameplay.

\section{MOBA Phases' Challenges}
\label{sec:phases}

A MOBA game can be divided in different phases, which one with specific challenges. Identifying such challenges and game features is important to better understand the game and to the discuss the main avenues for AI research on MOBA. In the next sections, for each phase, we describe its main characteristics and strategies and then present its main challenges in terms of AI. The phases are described in chronological order, starting before the match actually starts up to the point where the game state is more consolidated.

\subsection{Pick and Ban Phase}

\begin{figure}[h]
\begin{center}
\includegraphics[scale=0.20]{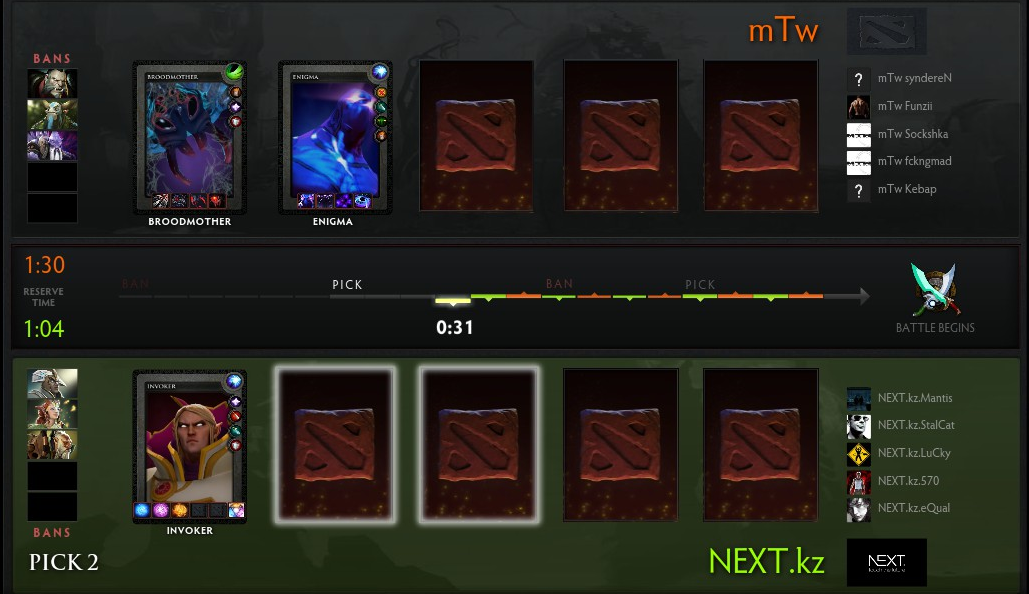}
\caption{A screenshot from the pick and ban Phase from the game Dota 2 in tournament mode.}
\label{fig:pickban}
\end{center}
\end{figure}

Most MOBAs offer a large hero base to players, allowing them to build specific strategies, and to pick heroes that excel over other heroes, an act called {\em counter-picking}. Normally, there are two main hero picking modes: {\em blind} and {\em tournament}. In blind mode each team of players selects freely from the hero pool without knowing the enemy team's heroes. On the other hand, in the tournament mode, each team picks alternately and is allowed to perform bans on the other team, as seen in Figure \ref{fig:pickban}. Moreover, in the tournament mode, the hero picked or being considered is seen by all players in the match. A hero cannot be picked twice in the tournament mode, while in blind mode it can be picked once by each team.

Most professional players and eSport staff agree that the game can be, sometimes, defined during the pick and ban phase. That is because there are some heroes that stand out in the game strategy, especially in professional gameplay where players have better mechanical skills. It is also common to see bans towards specific heroes that players have difficulty to play against, or that are considered {\em overpowered}.

The picks are normally made in accordance to a strategy to be executed by the team, such as fighting, picking enemies, pushing lanes, etc. The picks can also be done based on the player's skills with a specific hero or considering the {\em metagame}. The metagame consists of a set knowledge from inside and outside the game that encompass a series of guidelines  of which strategies and heroes are more likely to succeed during a time period. Lastly, the picks can be done based on outplaying the enemy team's strategy, or simply by picking heroes that counter-play the enemy heroes.

The choices in the Pick and Ban Phase are key since they can determine the outcome of the game. So, the challenges faced by an Artificial Intelligence Agent playing this phase are of great importance. Firstly, an Artificial Intelligent Agent should pick heroes that are effective both playing alone and in a team also considering that they should be at least as good as the other team's roster. Moreover, the AI should start the construction of a game strategy, planning which are the best actions for the chosen heroes. Lastly, it is important to perform some kind of opponent modeling in order to identify the possible strategies and roles to be assumed by the enemy heroes.

\subsection{Opening Phase}

The opening phase begin when the match starts and ends when the creeps or minions reach the middle of each lane. This phase consists of two main tasks: buying the opening items and performing role allocation. Optionally, there may be {\em pickoffs}, level one {\em team fights} or {\em jungle invasion}.

The act of buying the very first items  depends much on the enemies that the hero will be facing. Nevertheless, the initial items are very tentative, because the player can only guess which hero it will be playing against. If the player has a deep comprehension of the metagame, it can predict which hero will be in each lane, however, this can be just a guess. 

The role allocation not only consists of assigning a role to a hero, it also consists of making the initial positioning on the game environment. For instance, the strategy defined can propose an inversion of lanes to make the hero to face a weaker hero and, therefore, maximize its chances of winning the combat.

The jungle invasion consists of trying to kill neutral creeps located at the enemy's jungle, denying gold and experience to the enemy team, especially if there is a {\em jungle role}. Performing this strategy can lead to pickoffs, where the team finds a vulnerable hero in the map and try to kill him, getting gold and experience to the team. If the strategy is based on fights and the opportunity happens, the team can choose to perform a level one team fight, that happens before the heroes are well developed.

The main challenges here are the following: a) select the initial set of items to buy; b) allocate a set of five heroes to their respective roles; and c) define which lane will be claimed by one or more heroes. For accomplishing these, it may be necessary to develop a system that predicts the enemy's strategy and plan a counter strategy that minimizes the chances of losing the game, by defining roles, lanes and items to each hero.


\subsection{Laning Phase}

The Laning Phase starts  when the creeps reach the middle of each lane. It consists of mostly staying in the lane and trying to collect the maximum number of resources as possible. These resources come mainly from executing last-hits on enemy minions that come to the lane. This is generaly know as {\em farming}. Although this task seems to be simple, it is not, due to several factors.  The player competes in damaging with its allied creeps, allied heroes and turrets. Moreover, the player should take care to avoid being damaged by the enemies. In some games, like Dota and Dota2, the enemy heroes can also damage their allied creeps, making it harder to the player to execute the last hit.

Besides the resource collection task, the player should worry about the positioning of the heroes. Enemies can take advantage of a bad positioning by increasing the inflicted damage or even {\em ganking} the player. The ganking action consists of being caught and killed by other heroes that are not currently combating the player. A common practice is to use the Fog of War to perform effective ganks, taking advantage of the partial information. Every time the hero is killed the player receives a time punishment, through which the player cannot perform actions. This gank often comes from a player role called {\em ganker} or {\em jungler}, a role intended to help heroes to kill enemy heroes.

While most players act in the lanes, the jungler spends the laning phase collecting resources in the jungle. The jungler also roams around the map looking for opportunities to gank in the lanes. Moreover, junglers are responsible for the collection of minions and special blesses that change the hero status temporarily, commonly called buffs. For instance, in the game League of Legends, the jungle creeps offer special buffs that helps the jungler to perform ganks or help the team. In Dota2 and League of Legends there are heroes that can recruit neutral creeps to fight alongside them.

The Laning Phase normally finishes when a hero has enough resources to buy an important item or has relevant status or level difference to other heroes. That can be achieved through farming or killing enemies. Having this strategic advantage leads the hero to perform ganks itself performing the {\em roaming}, {\em i.e.} walking around the map in the jungle or other lanes. 

During the laning phase the AI should perform many reasoning tasks. Among these are: deciding whether or not to buy new items, when to fight or damage the enemy hero, which positioning is the best, etc. Moreover, since there are just three lanes and five heroes, the AI should be ready to cooperate with other players towards killing enemies or any other strategy.

\subsection{Mid Game}

The Mid Game is considered the phase in which the strategy takes form after the Laning Phase. Commonly there are three main strategy types found in most MOBA games: team fights, pickoff and lane push.

The team fight strategy consists of focusing the game in getting strategic advantage through defeating the enemy team in a series of group fights. Teams that focus in this strategy commonly have {\em carry} and {\em tank} heroes. The first is a hero capable of inflicting heavy damage on the opponents while the second is able of taking a lot of damage.  By winning those fights the entire team obtains gold and experience, while punishing the enemy team with time penalties. 

Pickoff is the action of selecting a vulnerable or weak hero and trying to kill it. Normally, this strategy focuses on defeating the enemy by picking important heroes and killing them to obtain fight advantage or weakening a main hero in the group. That can grant advantage in team fights or lead the team to gank the enemy heroes one-by-one, opening way to victory.

Lastly, the lane push strategy has the objective of picking a lane and pushing it as hard as possible, destroying all the structures on the path to the base. This kind of strategy has many variants, with three as the most used: {\em split push}, {\em sieging} and {\em team push}. The split push consists of making a distraction to the enemy team while a hero can destroy the enemy structures on other lanes. Sieging consists of pushing the enemy team hard toward their base or turrets, focusing on destroying enemy structures instead of fighting. Finally, the team push is the least used and may be seem as a rush strategy. The entire team goes for a lane and try to destroy all the structures as quick as possible, avoiding direct fights with the enemy team.

Challenges found in this phase are very diverse, but the cooperation and real-time reasoning problems stand out. In this phase, regardless of the strategy being followed, the synergy between the team is the key. Furthermore, the mechanics demonstrated by the AI should excel the human mechanics. Prediction, skill evasion and kiting are very valuable mechanics in this phase, because players will be fighting each other in a faster pace than in previous phases. A shallow analysis demonstrates that, for instance, a level one hero commonly has a few different actions to perform against one or two enemy heroes. On the other hand, in the mid game the hero has to fight against five other heroes and has many actions and action combinations.

\subsection{Late Game}

The late game is the last phase of a MOBA match. It starts when almost all heroes are missing one or two items to finish their itemization. In this phase the heroes are strong and the number of possible actions large, making combats more complex. Being caught alone in the late game normally means to be defeated. Moreover, the time penalty for being killed is higher at the late game, impairing the teams during fight.

When heroes have filled their item slots, they might start an itemization changing. It  consists  in changing the less valuable items for better ones. This technique also helps the team to counter-play the enemy strategies by using more appropriate items.

In the late game it is common to have the heroes at their maximum potential. Heroes that are meant to be {\em tank} can absorb great amounts of damage, while {\em carry} can inflict a huge amount of damage. The duels, or fights of one versus one, are mechanically driven, meaning that the most skilled player will win the fight, considering that their powers are almost the same.

Sieging and picking off heroes show their advantage at this time, since heroes can take turrets down faster. The split pushing is also very effective, but may be a weak strategy if the team has to temporarily fight without the member that is performing the split push. 

Finally, performing a combat outcome analysis at this level may be difficult, since there are many combinations of abilities and actions. Moreover, it is hard to predict the enemies actions, especially if the enemies are humans.

The challenges found here are mainly due to the huge amount of actions and  possible combinations. Besides, all the analysis should be performed in real-time, meaning that the reaction time is an important factor for the AI system. Generating an agent that is capable of reacting as fast as human players and capable of performing prediction and analysis at professional level is hard, because of the number of factors that impact these actions.

\subsection{Side Challenges}
\label{sec:side}

Along with the specific challenges found on each game phase, there are some challenges that are inherent to the game as a whole. These challenges are related with two types of analyses found in MOBA: prediction and combat analysis. These two fields converge in some sense, that the combat analysis is prediction dependent and vice-versa.

Here we list some of the prediction and combat challenges found in MOBA:
\begin{itemize}
    \item Combat Outcome from duels (1 on 1) and team fights;
    \item Killable enemies from a set of actions or combo;
    \item Strategies to be performed by the enemy team;
    \item Winner of the phase.
\end{itemize}

Regarding the combat analysis, the combat outcome prediction can only be performed by the analysis of the interactions between a group of heroes involved in the combat scenario. These interactions were previously studied in Dota 2~\cite{yang2014}. A special case of this scenario is the Duel, where two opposing heroes fight each other. Although seems to be easy, to predict the winner of a duel is hard, because it involves the prediction of attacks, abilities, movement, items, probabilities of critical damage, among others. Besides, the reaction in a duel should be done in real time, a task that requires the AI to return its answers as fast as possible. 

Concerning the field of counter playing the enemy team, it is important to predict which items the enemy proposes to buy. A successful build will help the hero to outplay the enemy by having statuses that help him to make the enemy hero's status less relevant. Moreover, building items in a non-static way is a hard task, because items should be acquired on demand, analyzing the match state. For instance, if the team is losing, one may prefer to buy defensive items, in order to have more survivability or to risk and buy items that maximize its damage, trying to be mechanically perfect. These challenges could be addressed by techniques that solve task allocation problems and optimization, like Ant Colony Optimization, Genetic Algorithms, among others.

\section{MOBA Mechanic Challenges}
\label{sec:mechanic}

The MOBA gameplay has various structures and mechanics that are common in most MOBA games. One of such structures are the turrets. The turret is a strong structure that guards the lane from being naturally advanced by the creeps. Besides, turrets provide a strong defensive area where heroes can stay safer than in the lane itself. The damage and behavior of the turrets provide a mechanical challenge common to the MOBA genre: the tower diving. The tower diving challenge is to go under the area of damage of the turret and kill one or more enemy heroes, leaving the dangerous area alive. 

Executing a tower diving action correctly, or even predicting its success, is challenging for both humans and Computers because it requires the analysis of a set of actions and preconditions. For example, the tower diving hero must be sure that it has sufficient HP to kill the enemy and is capable of receiving some hits from the turret without being killed. Moreover, there must be an analysis of  abilities, turret damage, probability of another enemy come to save the allied under turret are some of these features. One hero could simply stun the enemy diver under the turret area, leaving him there, stunned to death by turret damage. Both of those problems could be tackled using Influence Maps, Graphs Interaction analysis~\cite{yang2014} and Bayesian Networks. 

Another challenge that is common in the MOBA scenario is the Target Selection. Selecting the target in a fight is a hard task, especially in a cooperative mode. Humans, for instance, discuss with each other the enemy that should be the next target in their strategy. In case of AI controlled agents, this could be done by a voting or consensus algorithm, in which the agents would consider factors like low HP, high risk, strategy breaking, easy to kill, among others to decide the target.

In the laning phase, the hero should decide which minion should be the next target among a set of creeps called wave. In this phase it is common to perform the last hitting, which consists of hitting just the creeps that are killable by a single attack. On the other hand, the strategy could require the agents to push the lane, thus, the agents should hit continually, while maximizing the farm. This continuous hitting could be guided under rules that distribute the hits using a Target Selection Strategy.

The Target Selection should be coupled with a good positioning strategy, or this technique may not perform well. A good positioning should allow the hero to maximize its damage/effectiveness in the action performed, while minimizing the damage/disable received. An example of such positioning is to have long ranged units in the back, assuming that these units are easily killable but deal heavy damage. In the front line is interesting to position the melee and tank heroes, because they can absorb a lot of damage and retain the enemy advance, while their allies can inflict damage.

The previous discussion about positioning raises another relevant question. One may plan its strategy in order to minimize the damage received, however there are heroes that are {\em meant} to be damaged. Some of these heroes are initiators, and may not start the team fights without receiving damage from enemies. Moreover, there are heroes that take advantage of being damaged, like damage buffs, and even items that buffs the hero based on its missing HP. Therefore, the {\em damage minimization strategy} is not always the best strategy, and that should be taken in account in the agent development.

\section{A Discretized Model for MOBA Study}
\label{sec:discrete}

After the study of the MOBA structure, phases and elements, we see that the game is based on discrete parts, that can be isolated in order to perform a better study. Thus, we propose a model based on the individual core gameplay of each mechanic task from the MOBA game. Moreover, a testbed framework could be generated based on this model, making it possible to perform preliminary learning and testing in a faster pace.



Besides the testing benefit, it is known that MOBA, like its predecessor RTS, is divided in abstract decision layers. These layers mean that the agent should perform decision making on micro and macro levels. These abstract levels allow the AI to work in an individual skill without considering another ones. Moreover, the model is proposed in a way that each have no dependency on another part, simulating most or all features involved in each individual piece.

We propose the division of MOBA by the main tasks that need to be performed by an agent. The first task is the item build order, following, we propose models for farming, team fight, structure conquering and jungle. Having those models in an abstract form allows the research to make and test agents in diverse skills individually, further inserting this agent in a real game scenario.

\subsection{Item building}

As presented, the itemization is the process of buying items to improve the hero’s status. The validation of this process is commonly done in two ways: a) effectiveness of the hero in the game; and b) status increase. It is important to stress that the effectiveness in the game leads to status increase, since it provides more gold and experience to the hero and the team. On the other hand, status increase not always lead to effectiveness, since the agent must be mechanically perfect, which includes many variables and comprehension of the team behavior during cooperative tasks.

An effective agent should balance the items and be capable of reasoning which building order should be done in specific cases. This can be done without the need of simulating team fights, because the agent must only receive the enemy team's status and strategy and reason about which items should be bought to play against. This process can be done with static and dynamic planning, Reinforcement Learning, Neural Networks, among other techniques.

The item building model, therefore, must be composed of all items available in the game and should provide a reasonable amount of gold relatively to the phase to be analyzed. For instance, a little amount of gold should be provided to the agent to perform the openings, while a greater amount of gold should be provided for the mid or late game representations. The model should also be able to represent the different hero levels and possible enemy scenarios as inputs. It is clear that this is a hard task due to the large number of enemy team's combinations, enemy items, levels and game scenarios.

\subsection{Laning}

The laning phase is one of the most important phases in the MOBA gameplay. Winning this phase may imply in an important strategic advantage, besides more gold and experience. Learning how to lane is important for any AI agent, and is a difficult task since it involves many agents and structures.

The creation of a laning model is intended to help the agent to analyze the resource collection, targeting, harass and tower diving strategies. Therefore, it is necessary to implement a complete lane system, composed of heroes, creeps and turrets. The simulation/modeling of enemies is optional, since the agent could learn how to fight in another scenario, like the team-fight model.

This system could simulate a broad range of micromanagement skills. Learning how to control the lane, when to execute last hits, farming, and switching between advancing and retreating the creeps in the lane are essential skills to MOBA gameplay. In adversarial real-time gameplay, the system should be able to simulate multiple agents in the lane in order to show the harass and competition between agents for the creeps in the lane.

\subsection{Team fights}

The team fight model should allow the agents to interact with each other, and could be contained in a one lane map. This is intended to simulate the team fights and evolve the system to properly reason about combat results and possible interactions between agents. That system could be modelled both in a theoretical aspect as well as in a practical approach. 

The proposition of theoretical fight simulation is possible because the team fight consists of a series of interactions that is contained in a timespan. That could be modelled as a temporal graph, pointing the interactions between the agents. In fact, professional players and coaches already do that simulation which they call {\em theorycraft}, which consists of a mathematical analysis of the game~\cite{paul2011}. After performing this analysis, the player formulates theories of which interactions and items give most advantage to his game strategy/hero. This field could take benefit of models already established like the {\em econometrics}~\cite{judge1988}.

Many combinations and analysis could be made in an effective team fight model. Among these analysis is the hero balancing in the MOBA environment, a task that is done by a game designer team. Given the MOBA ecosystem, we observe that the game nature is clearly evolutionary, given its constant updates. Moreover, the combat outcome and the potential of a hero during all game phases could be analyzed, determining interesting strategies, from duels to 5v5 team fights.

\subsection{A Theoretical MOBA model}

\begin{figure}[h]
\begin{center}
\includegraphics[scale=0.18]{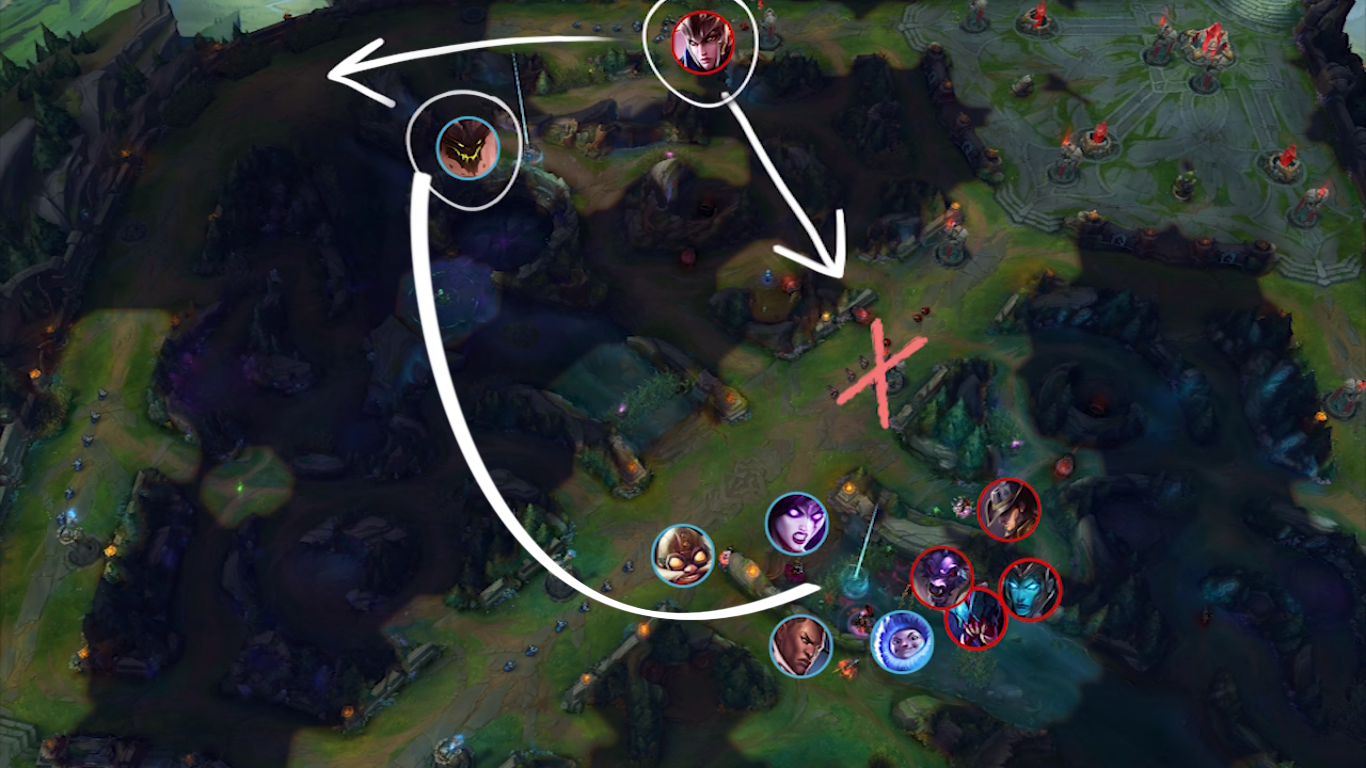}
\caption{Example of a strategic analysis in a MOBA match. The top laners take different approaches for helping their teams.}
\label{fig:boardanalysis}
\end{center}
\end{figure}

Besides the discretization of the MOBA game, the game as a whole can be analyzed using a theoretical model, as done in most sports such as soccer, football and rugby. This model could be also considered in theorycraft by the coaches and the players while planning the positioning, hero groups and strategies. An example is shown in Figure \ref{fig:boardanalysis}. 

It may be faster to simulate games using graphs and theoretical models in order to predict or analyze strategic situations and positions. This analysis can even consider the probabilities of succeeding in the execution of game mechanics, being modelled as a probabilistic graphical model. Such probabilities may be computed from the performance obtained in the previous discrete parts, and should be identified in the strategy and game style adopted by a team and its strategy.

Such model could even be used to identify emerging strategies that are unusual to players and teams, outperforming their common strategies. Lastly, modeling replays in a theoretical representation can help the system to classify the strategies, identifying their main characteristics. For instance, distinguishing split push from sieging, or team fight strategy from pickoff. This classification can help the system to map the strategies adopted by a team or a hero group.

\begin{figure}[h]
\begin{center}
\includegraphics[scale=0.26]{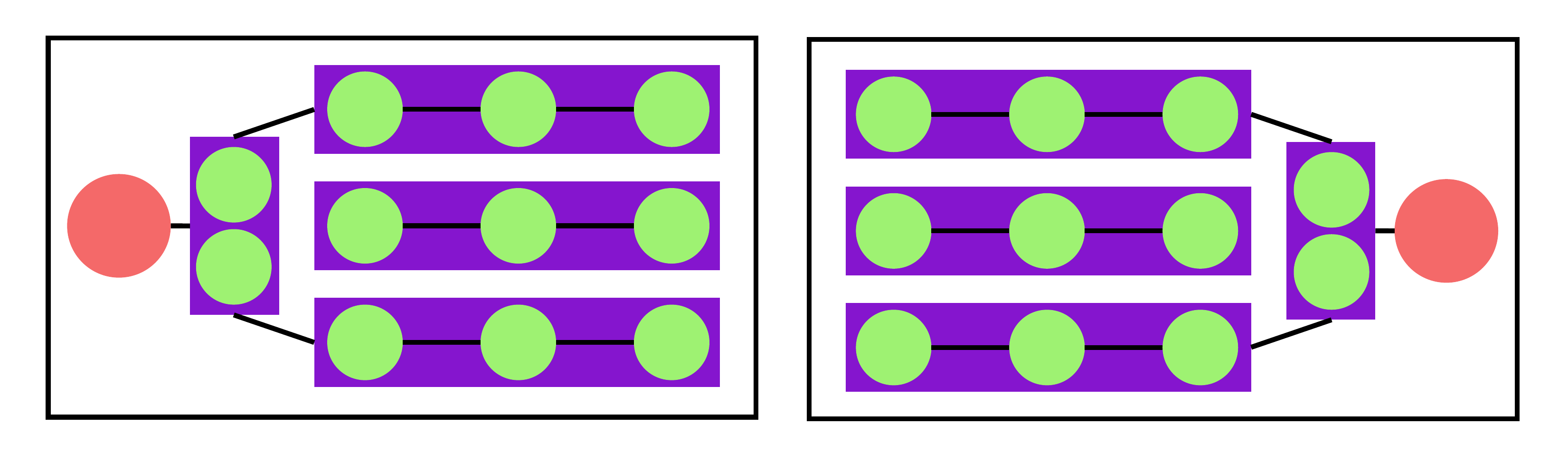}
\caption{Diagram of a theoretical structure to represent a MOBA game. The smaller nodes represent turrets while the larger one is the main structure.}
\label{fig:MOBADiagram}
\end{center}
\end{figure}

The model follows the conventions established by the MOBA map, with its structures and lanes. We translate the structures into nodes and the order of destruction of these structure into edges. As discussed, the outermost structure should be destroyed in order to enable the damage or destruction of an inner structure. Finally, the structures of an entire lane should be destroyed in order to enable the damage to the main structure. A diagram showing the graph structure of the model can be seen in Figure \ref{fig:MOBADiagram}.

Notice in the diagram that the interactions start in the outermost structure, meaning that the turrets next to the center should be destroyed first. 
The rectangles containing a sub-graph represent a dependency among the structures, such that structures in different rectangles are independent from each other. Destroying a group of structures unlocks a way to attack the two turrets protecting the main structure. Lastly, after destroying these turrets, the team can attack the main structure. Its eventual destruction causes the end of the game.

\section{Related Work}
\label{sec:related}

Despite the popularity of the genre, MOBA has not attracted much attention from researchers. This lack of attention could be mainly attributed to the commercial nature of these games and the lack of collaboration of industry and academia, in terms of open software, APIs, etc. Besides, MOBA is a relatively new game genre, with the main titles debuting in the late 2000's. 

The MOBA history go back to the RTS hype, when modders used to modify RTS maps in Starcraft and Warcraft III to make unique gameplay. The RTS inclusively got much attention from the academia, having been widely studied and used as a competition platform for many years \cite{buro2003,ontanon2013}. In that platform, we saw the use of Starcraft as a testbed for many AI techniques, and the development of a complete agent capable of defeating human players is still an open problem. Problems like unit management \cite{synnaeve2011}, tactical gameplay \cite{hagelback2012}, kiting \cite{uriarte2012}, among others, were extensively studied in the RTS environment.

Some of these problems have also been studied in the MOBA environment. One example is the Kiting, a mechanic to hit an run, largely used in the MOBA gameplay. This problem was tackled before Uriarte and Onta\~n\'on~\cite{uriarte2012} using an ad-hoc approach, where they model the units involved in the Target Selection task using multiple features and status analysis, which is commonly called aggro. Moreover, selecting the target is not only related to hero versus hero, but also in the farming action.

Later, researchers started to note that MOBA brings a different type of challenge. Instead of challenging the player to manage large armies or to optimize build orders, MOBA invites the player to control a single unit and master it.

The popularity of MOBA expanded so fast that three years after its debut it was already the most played genre in the world, surpassing the Multiplayer Massively Online Real Playing Game, World of Warcraft. That expansion got the attention of some researchers that wanted to better understand the MOBA dynamics. 

The first researchers were worried to present to the readers the MOBA phenomena, its basic characteristics and gameplay. In \cite{nosrati2012} and \cite{rioult2014} we can find information about basic MOBA characteristics. A broad discussion of MOBA as an eSport and game design can be found in \cite{ferrari2013}.

After these descriptive works, some researchers wanted to go inside the MOBA gameplay and analyze it from the players perspective or to observe player's performance. Because there were no tools to do so, most works have taken advantage of large replay databases to develop their work. The behavior of MOBA players during a match and its impacts into winning the game are explored in~\cite{pobiedina2013}. Later, the analysis of MOBA player’s performance is done in \cite{drachen2014}, using a \textsf{Dota2} replay database as input. On the other hand, the work of \cite{yang2014} is interested in showing that MOBA has some patterns that could lead a team to victory. It is interesting to see that both works mention that the knowledge developed could be used to create intelligent agents for MOBAs, which happened later.

The urge of creating agents for MOBA games was partially satisfied with the tools available by S2Games in their game \textsf{Heroes of Newerth}. However, some users got frustrated because there was not much information about the API, besides the need of using some preprogrammed behaviors. In that context we find the work of \cite{will2015}, which used Reinforcement Learning techniques for the implementation of a MOBA agent. Later, we find the work of \cite{nascimento2015a} that uses third part tools to develop agents in the League of Legends context.


\section{Conclusions and Future Work}
\label{sec:conclusion}

In this paper we presented an overview of the Multiplayer Online Battle Arena genre as a testbed for Game AI research. We discuss the challenges introduced in each phase of the game and also in the game as a whole and propose a model that can be implemented to allow a better understanding of the game.

We observe that the research interest in MOBA games is increasing, but it is still incipient. By presenting and discussing the main features and challenges introduced by MOBA games, we aim to ``push the lane" into the direction of a better understanding and increased use of MOBA games as a research platform.



For future work we believe that the discrete framework proposed in this paper should be developed in order to provide a reliable testbed for research.  Moreover, the knowledge extracted from that framework could be useful in real MOBA games, like Smite, Dota2 or in the RTS Starcraft. 

\bibliographystyle{IEEEtran}
\bibliography{bare_conf}

\end{document}